\documentclass{article}

\usepackage[preprint]{neurips_2026}

\usepackage[utf8]{inputenc}
\usepackage[T1]{fontenc}
\usepackage{hyperref}
\usepackage{url}
\usepackage{booktabs}
\usepackage{amsfonts}
\usepackage{amsmath}
\usepackage{amssymb}
\usepackage{nicefrac}
\usepackage{microtype}
\usepackage{xcolor}
\usepackage{graphicx}
\usepackage{multirow}
\usepackage{caption}
\usepackage{algorithm}
\usepackage{algpseudocode}
\usepackage{cleveref}
\crefname{appendix}{Appendix}{Appendices}
\Crefname{appendix}{Appendix}{Appendices}

\newcommand{\E}{\mathbb{E}}
\newcommand{\V}{\mathcal{V}}

\DeclareMathOperator{\KLop}{KL}

\newcommand{\teacher}{\mathbf{p}}  % the teacher distribution; reserve subscript/superscript for caller
\newcommand{\student}{\pi_\theta}
\newcommand{\opsdtarget}{\mathbf{p}^{\mathrm{opsd}}}

\newcommand{\precision}{\kappa}

\newcommand{\opsd}{\textsc{OPSD}}
\newcommand{\eopd}{\textsc{EOPD}}
\newcommand{\todi}{\textsc{ToDi}}

\newcommand{\pwopsd}{\textsc{PW-OPSD}}

\title{When Are Teacher Tokens Reliable?\\
Position-Weighted On-Policy Self-Distillation for Reasoning}

\author{%
  Xiaogeng Liu\textsuperscript{1}\thanks{Corresponding to \texttt{xliu316@jhu.edu}.} \quad
  Xinyan Wang\textsuperscript{2} \quad
  Yingzi Ma\textsuperscript{2} \quad
  Yechao Zhang\textsuperscript{3} \quad
  Chaowei Xiao\textsuperscript{1} \\[0.6em]
  \normalsize
  \textsuperscript{1}Johns Hopkins University \quad
  \textsuperscript{2}University of Wisconsin--Madison \\
  \textsuperscript{3}Nanyang Technological University
}

\begin{document}

\maketitle

\begin{abstract}
On-policy self-distillation (\opsd{}) trains a student on its own rollouts
using a privileged teacher, but its standard objective weights all generated
tokens equally, implicitly treating the privileged teacher target as equally
reliable at every student-visited prefix. Existing entropy-based OPD methods
relax this uniformity by modulating token-level supervision with teacher
entropy, but high teacher entropy in reasoning has an ambiguous reliability
meaning: it can reflect either non-viable uncertainty or benign solution
diversity. To identify this phenomenon, we introduce a branch-viability diagnostic.
Specifically, we record next-token alternatives from the
privileged-answer teacher prompt, force each alternative after the student
prompt plus its on-policy spine prefix, and test whether the resulting
student-template continuation recovers the correct answer. On
Qwen3-4B, we find that an oriented within-sequence position score is the strongest tested
predictor of teacher-token reliability,
reaching an area-under-ROC-curve (AUROC) of $0.83$ with a $95\%$
cluster-bootstrap interval of $[0.66,0.95]$; local uncertainty scores are at
most $0.57$. Motivated by this trajectory-level structure,
we propose Position-Weighted On-Policy Self-Distillation (\pwopsd{}), which applies an
increasing position weight while keeping the same student rollout, privileged
teacher pass, and clipped forward-KL target as \opsd{}. In our comprehensive evaluations with different random seeds, the diagnostic-derived \pwopsd{} improves AIME 2024
and AIME 2025 Avg@12 by $+1.0$ and $+1.1$ points, and a generalization evaluation on two larger-scale models from different families, DeepSeek-R1-Distill-Llama-8B and Olmo-3-7B-Think, also demonstrates consistent aggregate Avg@12 improvements. These results show that teacher-token reliability in reasoning
distillation is trajectory-structured and can be utilized without additional
teacher computation. The code is available at \url{https://github.com/SaFo-Lab/PW-OPSD}
\end{abstract}

% Deleted " matches \opsd{} on MATH-500 Avg@12 ($95.34$ vs.\ $95.33$) and " and "; a stronger schedule from the sweep improves HMMT 2025 Avg@12 by $+1.48$ points over \opsd{}"

\section{Introduction}
\label{sec:intro}

On-policy self-distillation (\opsd{})~\citep{zhao2026opsd} is a practical
recipe for distilling mathematical reasoning because it supervises the student
on prefixes the student actually visits. The student samples a rollout from
the ordinary problem prompt, and a privileged copy of the model, conditioned
on reference information, provides token-level targets along that same
rollout. This construction avoids the trajectory mismatch of teacher-generated
demonstrations while still injecting privileged information at training time.

The standard \opsd{} objective nevertheless makes an implicit reliability
assumption: every generated token receives the same forward-KL
supervision~\citep{zhao2026opsd}. This uniform objective treats the privileged
teacher target as equally useful at every student-visited prefix. Recent
adaptive, relaxed, or gated distillation objectives have challenged this
uniformity assumption from complementary angles: \eopd{} augments reverse-KL
OPD with forward KL on high-teacher-entropy tokens~\citep{jin2026eopd},
\todi{} mixes forward and reverse KL per token using a teacher-student
probability log-ratio~\citep{jung2025todi}, REOPOLD treats the
teacher-student log-likelihood ratio as a token reward with reward clipping
and entropy-based sampling~\citep{ko2026scalingreasoningefficientlyrelaxed},
and GATES gates privileged self-distillation by tutor consensus~\citep{stein2026gates}.
Together, these works motivate nonuniform token-level supervision through
local entropy, teacher-student mismatch, clipped token rewards, or consensus
signals.
\begin{figure}[t]
\centering
\includegraphics[width=0.9\linewidth]{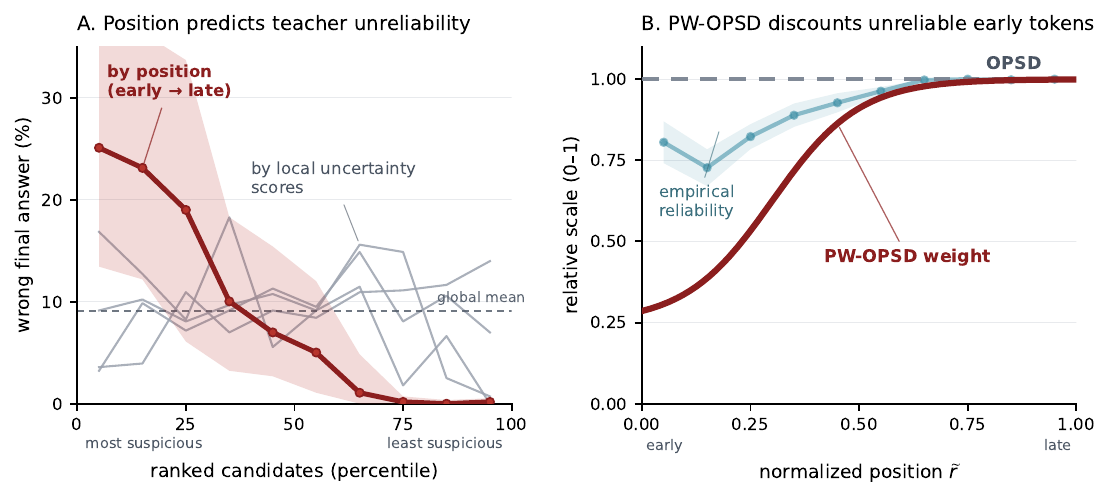}
\vspace{-0.2cm}
\caption{Branch viability reveals a positional reliability structure in
Qwen3-4B reasoning traces. Alternatives are selected by the same model
under the privileged-answer teacher prompt and rolled out under the
ordinary student prompt; for high-ambiguity candidate positions, these
forced student-context continuations fail more often at early positions
and rarely fail late in the trace. \textbf{(A)} Sorting candidates by
normalized position therefore separates teacher-unreliable branch points
from viable diversity, whereas local uncertainty (entropy) scores yield
near-flat failure curves. \textbf{(B)} The resulting empirical reliability
curve motivates the \pwopsd{} weight schedule, which discounts early
high-ambiguity positions while preserving full-strength supervision later.
See \cref{sec:method:motivation} for the details of the diagnostic
experiments.}
\vspace{-0.5cm}
\label{fig:branch-auroc}
\end{figure}
However, all of these adaptive criteria share a common limitation: they
measure how ambiguous the teacher's local distribution is, through entropy,
teacher-student mismatch, or tutor consensus, but token-local ambiguity
is \emph{not} the same as low teacher reliability, and conflating the two
reweights tokens for the wrong reason. Our point of departure is the
reliability meaning of these local signals.
Entropy-based criteria are a natural way to detect uncertainty, but high
teacher entropy is not the same as low teacher reliability. A high-entropy
teacher distribution can reflect non-viable uncertainty: several locally plausible
next-token alternatives receive probability mass, but some of them fail to
recover the correct answer under continued student-context generation. However, it can
also reflect benign diversity: the teacher assigns mass to multiple viable
solution routes or surface realizations that still preserve correctness.
Thus a token-local uncertainty score can identify ambiguity, but it does not
distinguish non-viable uncertainty from viable solution diversity.

The practical cost is that any reweighting following local ambiguity alone
will either amplify supervision at branch points where the teacher derails,
or downweight supervision at benign branch points where the teacher stays
correct. Thus, the right object to estimate for adaptive token weighting is
not teacher entropy but teacher \emph{reliability}: a per-context probability
that the teacher's local target is useful supervision for the student. We formalize this \emph{reliability} as a latent indicator $I_t$ for
whether matching the teacher target at distillation context $c_t$ helps
preserve or recover the correct solution. The reliability-weighted
surrogate then weights the forward-KL term by
$\rho^*(c_t)=\Pr(I_t=1\mid c_t)$. 
% This view identifies the object that adaptive
% distillation should estimate: not teacher entropy per se, but the probability
% that the teacher's local target is useful for the student at that context.

To connect this latent quantity to observable behavior, we introduce a
branch-viability diagnostic. Throughout, ``teacher'' and ``student'' refer
to two prompt templates applied to the same model: the teacher
template includes the privileged ground-truth answer; the student template
is the ordinary problem prompt. Starting from a student-template rollout
that reaches the correct answer, we use the teacher template to propose
high-ambiguity next-token alternatives, force each alternative after the
student-template prompt plus the spine prefix, and continue under the
student template (no privileged information). We use student-template
continuation because a teacher-template continuation can re-use the
privileged answer and recover from almost any forced branch, collapsing
the labels. If the forced alternatives usually
fail to recover the correct answer, the candidate is labeled
\emph{real-uncertain}: at that prefix, the teacher's local alternatives are
low-reliability supervision for the student distribution. If the alternatives
remain viable, the ambiguity is treated as benign diversity rather than as
evidence of low teacher reliability.

On Qwen3-4B~\citep{qwen3}, this diagnostic shows a clear positional
pattern, as demonstrated in Fig.~\ref{fig:branch-auroc}. In the correct-spine
subset, real-uncertain candidates concentrate early in the normalized trace,
while diversity candidates are more common later. After
within-problem residualization, an oriented position score separates
real-uncertain candidates from diversity candidates with AUROC $0.83$ and a
$95\%$ cluster-bootstrap interval of $[0.66,0.95]$; the tested local
uncertainty diagnostics reach residual AUROC at most $0.57$. Thus the
diagnostic supplies the direction of the reliability prior: early
high-ambiguity branch points are where the privileged teacher's local
alternatives most often become unreliable, while later positions are more
reliable under forced continuation.

Inspired by this finding, we propose Position-Weighted On-Policy Self-Distillation (\pwopsd{}), which implements this
reliability trend as a plug-in approximation to $\rho^*(c_t)$. It keeps the
same student rollout, privileged teacher pass, and per-vocabulary clipped
forward-KL surrogate as \opsd{}, but aggregates token losses with an increasing
sigmoid function of normalized within-sequence position. The floor keeps early
tokens partially supervised, while the increasing schedule assigns stronger
weight to later positions where the diagnostic indicates higher teacher
reliability. The method therefore changes only the outer reliability-weighted
aggregation, adding no extra teacher pass or auxiliary verifier.

We evaluate \pwopsd{} on Qwen3-4B math reasoning benchmarks with the same
maximum generation length of $38{,}912$ tokens as the \opsd{} reference
evaluation~\citep{zhao2026opsd}. The
diagnostic-derived schedule matches \opsd{} on MATH-500 Avg@12
($95.34$ vs.\ $95.33$) and improves Avg@12 by $+1.0$ point on AIME 2024 and
$+1.1$ points on AIME 2025, with the same teacher cost. The position-schedule
sweep further shows how reliability-curve strength affects harder regimes:
the aggressive schedule improves HMMT 2025 Avg@12 by $+1.48$ points over
\opsd{} and improves Maj@12 by $+1.11$ points. These results support position as a structural reliability prior and
show that reasoning distillation benefits from modeling teacher reliability
along the trajectory.

% To summarize, we make three contributions.
% \begin{itemize}\itemsep=0pt
% \vspace{-0.2cm}
% \item[(1)] \textbf{A branch-viability diagnostic.} We measure whether
% teacher-preferred alternatives at high-ambiguity positions remain viable under
% student-context continuation, separating non-viable uncertainty from benign
% diversity.
% \item[(2)] \textbf{A position-weighted OPSD objective.} \pwopsd{} applies an
% increasing sigmoid reliability weight over normalized within-sequence position
% while preserving the rollout, teacher pass, and clipped forward-KL target of
% \opsd{}.
% \item[(3)] \textbf{An empirical evaluation.} On Qwen3-4B, \pwopsd{} improves
% AIME 2024 and AIME 2025 Avg@12, matches MATH-500, and uses schedule
% sensitivity on HMMT 2025 to identify harder regimes where the reliability
% curve should be strengthened or learned.
% \end{itemize}

\section{Related Work}
\label{sec:related}

\paragraph{On-policy distillation.}
Knowledge distillation trains a student to match soft teacher probabilities
rather than only hard labels~\citep{hinton2015distilling}. Sequence-level
distillation extends this idea from token labels to generated completions
\citep{kim2016sequence}. Modern LLM post-training popularized instruction
and feedback-based tuning~\citep{ouyang2022training}; in distillation, recent
LLM methods often move supervision on policy, so the student is trained on
prefixes it actually visits~\citep{agarwal2024gkd,gu2024minillm,ko2024distillm,zhao2026opsd}.
This follows the interactive imitation-learning intuition of querying expert
feedback on learner-visited
states~\citep{ross2011reduction}; empirical work on LLM imitation also reports
degradation under off-policy imitation in multi-step generation settings
\citep{gudibande2023falsepromise}. GKD formalizes generalized on-policy
knowledge distillation~\citep{agarwal2024gkd}, MiniLLM optimizes
sequence-level reverse-KL distillation on student/teacher-mixed on-policy
samples through a policy-gradient formulation~\citep{gu2024minillm}, and
DistiLLM studies skewed KL objectives for white-box OPD on student-generated
outputs~\citep{ko2024distillm}. See \citet{song2026survey} for a broader survey
of OPD. \opsd{} is the main
baseline in this paper: it uses a
privileged teacher prompt containing reference information to provide
token-level targets along student rollouts~\citep{zhao2026opsd}. \pwopsd{}
keeps the privileged on-policy construction of \opsd{} and makes the outer
reliability weight trajectory-structured rather than uniform.

\paragraph{Divergence choice and adaptive token weighting.}
Several distillation methods change the divergence or token-level controller
between teacher and student. MiniLLM replaces standard forward-KL KD with
sequence-level reverse KL and optimizes the resulting on-policy objective via
policy-gradient returns~\citep{gu2024minillm}; DistiLLM uses skewed KL
objectives to stabilize white-box OPD on student-generated
outputs~\citep{ko2024distillm}. Broader $f$-divergence work
studies how divergence choice changes sequence-level KD behavior
\citep{wen2023fdivergence}. AKL revisits the usual mode-seeking/mode-covering
framing for LLM distillation, arguing that forward and reverse KL share the same
asymptotic objective but that FKL emphasizes head probabilities while RKL
emphasizes tail probabilities in early epochs, then proposing an adaptive KL
mixture~\citep{wu2024akl}. Recent adaptive or privileged distillation methods
also depart from uniform token-level supervision, but for different reasons and
with different control signals. \eopd{} applies forward KL on
high-teacher-entropy tokens while retaining reverse KL elsewhere in
OPD~\citep{jin2026eopd};
REOPOLD~\citep{ko2026scalingreasoningefficientlyrelaxed} interprets OPD as
policy optimization with a teacher-student log-likelihood-ratio token reward
and relaxes strict imitation through reward clipping, entropy-based dynamic
sampling, and exploration-to-refinement training. Outside this student-rollout
OPD divergence-switching line, \todi{} adaptively combines forward and reverse
KL per token using a teacher-student probability log-ratio~\citep{jung2025todi};
GATES instead gates privileged self-distillation using consensus among tutor
traces~\citep{stein2026gates}.
These methods show that distillation supervision can be modulated per token
through entropy, teacher-student ratios, clipped rewards, or consensus signals.
However, these local control signals do not directly distinguish non-viable
uncertainty from benign solution diversity. To explore this challenge, we
propose a branch-viability experiment in which the same model, under
the privileged-answer teacher prompt, selects forced alternatives and then
rolls them out under the ordinary student prompt to test whether the
correct answer is still recovered.
We compare branch viability against standard uncertainty
diagnostics, including predictive entropy, MC-dropout mutual information, and
Dirichlet or evidential uncertainty~\citep{kendall2017uncertainties,gal2016dropout,malinin2018predictive,sensoy2018evidential}.
These baselines quantify local ambiguity, but they do not test whether
high-probability teacher alternatives preserve correctness under continued
generation. 

\section{Method}
\label{sec:method}

\pwopsd{} is a reliability-weighted version of \opsd{} that preserves the
student rollout, privileged teacher pass, and per-vocabulary clipped
forward-KL surrogate. It changes the outer token aggregation through a
position-dependent reliability weight and a per-sequence mean, motivated by a
branch-viability diagnostic showing that early high-ambiguity branch points
are often teacher-unreliable, while local uncertainty scores are weak
predictors of this reliability event.

\subsection{Background: On-Policy Self-Distillation}
\label{sec:method:opsd}

Let $x_{\mathrm{stu}}$ be the ordinary student prompt and
$x_{\mathrm{tch}}$ be the privileged teacher prompt that includes the
reference solution. On-policy self-distillation
(\opsd{})~\citep{zhao2026opsd} first samples a completion from the current
student,
\[
y_{1:T} \sim \student(\cdot \mid x_{\mathrm{stu}}),
\]
and then queries a teacher copy of the same model on the privileged context
and the student prefix. At each valid generated position $t$, the teacher
target is
\[
\opsdtarget_t
= p_{\mathrm{tch}}(\cdot \mid x_{\mathrm{tch}}, y_{<t}),
\qquad
p_t
= \student(\cdot \mid x_{\mathrm{stu}}, y_{<t}).
\]
Both distributions are temperature-scaled before the softmax, matching the
trainer path used for all forward-KL losses. With valid-token set
$\mathcal{M}$, the implemented \opsd{} objective is
\begin{equation}
\label{eq:opsd}
\mathcal{L}_{\opsd{}}
=
\frac{1}{|\mathcal{M}|}
\sum_{t \in \mathcal{M}}
\sum_{j \in \V}
\min\!\left(
\opsdtarget_t(j)\log\frac{\opsdtarget_t(j)}{p_t(j)},
\tau_{\mathrm{clip}}
\right).
\end{equation}
The clipping in \cref{eq:opsd} is element-wise over vocabulary terms: the
quantity
$\opsdtarget_t(j)\log(\opsdtarget_t(j)/p_t(j))$ is clipped for each
$j \in \V$, the clipped terms are summed over the vocabulary, and the result
is averaged over valid tokens. This objective is the uniform-weight baseline
whose outer token aggregation \pwopsd{} will modify.

\subsection{Motivation: Token-Position Predicts Teacher Reliability}
\label{sec:method:motivation}

\paragraph{Branch-viability protocol.}
The branch-viability diagnostic asks whether a locally ambiguous teacher
preference is reliable supervision for the student distribution. Throughout
the diagnostic, ``teacher'' and ``student'' refer to two prompt templates
applied to the same Qwen3-4B model: the \emph{teacher} template
includes the privileged ground-truth answer, while the \emph{student}
template is the ordinary problem prompt without that information.

For each problem, the model under the student template first generates an
on-policy rollout; this fixed token sequence is the \emph{spine}. The model
is then re-evaluated along the spine under the teacher template, which adds
the privileged answer. Candidate positions are pre-selected from high
top-$16$ valid-token truncated entropy on this teacher pass;
Appendix \ref{app:branch-viability} gives the filtering details. The
diagnostic is deliberately targeted: it studies positions where the teacher
already signals local ambiguity, rather than arbitrary positions on the
rollout. The reliability-weighted view in \cref{sec:method:theory}
identifies the latent reliability posterior as the formal object that this
diagnostic probes behaviorally.

At each candidate position, the teacher proposes preferred next-token
alternatives. We then test whether each alternative leads back to the
correct answer \emph{from the student's perspective}: we form a prefix
consisting of the student-template prompt, the spine truncated to the
candidate position, and the forced alternative token, and complete the
sequence under the student template (no privileged information). A
candidate is labeled \emph{real-uncertain} when these student-context
continuations fail to recover the correct final answer; in this case
the teacher's local alternatives are unreliable targets for that student
prefix. A candidate is labeled \emph{diversity} when the alternatives
remain viable routes to the same answer. We continue under the student
template rather than the teacher template because student rollouts and
deployment use the ordinary prompt without privileged information, and a
teacher-template continuation can recover from almost any forced branch
by re-using the ground-truth answer in its prompt; that recovery
affordance would mask the very reliability question the diagnostic is
asking. Appendix
\ref{app:branch-viability} gives the thresholded implementation details
for the repeated forced-continuation rollouts.

\paragraph{Empirical finding.}
\Cref{tab:branch-auroc} reports $8$ real-uncertain and $271$ diversity
candidates from $61$ problems, restricted to the correct-spine subset; the
underlying failure-rate-by-position curves are visualized in
Fig.~\ref{fig:branch-auroc}. Each
feature is residualized within problem before scoring, so the AUROC reflects
within-problem token-level separation rather than problem difficulty.
Confidence intervals are cluster bootstraps over problems.

Position is the strongest predictor among the diagnostics we test. We define
normalized position as $\widetilde r = (\text{spine\_pos}+0.5)/L$, where $L$ is the length of the student spine. Since the real-uncertain
candidates concentrate early, \cref{tab:branch-auroc} reports the oriented
early-position score $1-\widetilde r$ for the real-uncertain label. This score
reaches AUROC $0.83$ with $95\%$ CI $[0.66,0.95]$, whereas predictive
entropy, MC-dropout mutual information, Dirichlet precision, and top-$16$
truncated entropy remain weak predictors. Equivalently, raw position
$\widetilde r$ is negatively associated with teacher unreliability, which is
why \pwopsd{} uses an increasing reliability weight.

\begin{table}[t]
\centering
\small
\caption{Branch-viability classification on Qwen3-4B. Entries are
within-problem residualized AUROC for separating $8$ \emph{real-uncertain}
candidates (whose forced student-context continuations fail to recover the
correct answer) from $271$ \emph{diversity} candidates (whose continuations
remain viable) across $61$ problems, restricted to the correct-spine subset
and to high-truncated-entropy candidate positions.
For position, the scored feature is $1-\widetilde r$, so larger values
correspond to earlier tokens. Cluster-bootstrap $95\%$ CIs are computed over
problems.}
\label{tab:branch-auroc}
\begin{tabular*}{0.95\linewidth}{@{\extracolsep{\fill}}lcc@{}}
\toprule
score (residualized within problem) & residual AUROC & 95\% CI\\
\midrule
$H(\opsdtarget_t)$ (mean predictive entropy) & $0.51$ & $[0.50, 0.84]$\\
$\mathrm{MI}$ (canonical MC-dropout epistemic) & $0.51$ & $[0.50, 0.62]$\\
$\log\hat{\precision}$ (Dirichlet precision) & $0.54$ & $[0.50, 0.69]$\\
$h_{\mathrm{trunc}}$ (top-$16$ valid entropy) & $0.57$ & $[0.51, 0.72]$\\
$1-\widetilde r$ (early-position score) & $\mathbf{0.83}$ & $\mathbf{[0.66, 0.95]}$\\
\bottomrule
\end{tabular*}
\vspace{-0.5cm}
\end{table}

The local diagnostics in \cref{tab:branch-auroc} include standard teacher-side
uncertainty scores: predictive entropy~\citep{kendall2017uncertainties},
MC-dropout mutual information~\citep{gal2016dropout}, and Dirichlet precision
and evidential-style categorical uncertainty~\citep{malinin2018predictive,sensoy2018evidential},
plus our top-$16$ truncated entropy. They are included as diagnostic baselines
for reliability prediction; implementation details for the MC-dropout scores
are in \cref{app:mc-dropout}.

\paragraph{Implication for adaptive distillation signals.}
The weak scores in \cref{tab:branch-auroc} call into question a common
assumption behind recent adaptive, relaxed, or privileged distillation methods:
that local uncertainty or teacher--student discrepancy can serve as a reliable
control signal for distillation. Existing methods instantiate this idea through
teacher token entropy in \eopd{}~\citep{jin2026eopd}, token-wise
teacher--student probability log-ratios in \todi{}~\citep{jung2025todi},
teacher--student log-likelihood-ratio rewards and entropy-guided token-level
sampling in REOPOLD~\citep{ko2026scalingreasoningefficientlyrelaxed}, and
tutor-answer consensus in GATES~\citep{stein2026gates}. These signals are useful
for adapting objectives, rewards, or trajectory selection, but our diagnostic
shows that they are not reliable proxies for branch viability: the same large or
permissive signal can arise when the teacher is unreliable and when multiple
continuations are genuinely viable. Position provides a complementary structural
signal, capturing where local ambiguity becomes unreliable for the student.

\subsection{Theoretical Interpretation: Reliability-Weighted Distillation}
\label{sec:method:theory}

Let $c_t=(x_{\mathrm{stu}},x_{\mathrm{tch}},y_{<t})$ denote the distillation
context, let $q_t$ be the privileged teacher target at that context, and let
$p_t=\student(\cdot\mid x_{\mathrm{stu}},y_{<t})$ be the student distribution.
For the interpretation, define the unclipped token divergence
$D_t(\theta)=\KLop(q_t\,\|\,p_t)$. Introduce a latent indicator
$I_t \in \{0,1\}$, where $I_t=1$ means that the teacher's local target is
reliable for the student prefix: matching it helps preserve or recover the
correct solution rather than following an off-path alternative. The ideal
reliability-filtered risk is
\begin{equation}
\label{eq:reliability-risk}
R(\theta)
=
\E\!\left[I_t \cdot D_t(\theta)\right].
\end{equation}
For fixed $\theta$, $D_t(\theta)$ is determined by the distillation context, so
the tower property gives
\begin{equation}
\label{eq:reliability-weight}
R(\theta)
=
\E\!\left[\rho^*(c_t) \cdot D_t(\theta)\right],
\qquad
\rho^*(c_t)=\Pr(I_t=1\mid c_t).
\end{equation}
The Bayes-optimal surrogate for this latent-reliability risk weights each
token by the posterior probability that the teacher target is reliable.
Appendix \ref{app:reliability-surrogate} gives the conditioning details.
On-policy and adaptive distillation methods exercise related levers through
sampling distributions, divergences, or weight
functionals~\citep{agarwal2024gkd,gu2024minillm,ko2024distillm,wen2023fdivergence,wu2024akl,jung2025todi,jin2026eopd};
\pwopsd{} uses normalized position as a low-cost structural proxy for that
posterior.

The branch issue can also be summarized by a mixture identity. Let
$h_t=(x_{\mathrm{stu}},y_{<t})$ denote the student-side prefix. Suppose the
teacher target at that prefix is a mixture over latent successful branches $Z$,
\[
q_t = \sum_z \alpha(z\mid h_t) q_t^z .
\]
For any student distribution $p_t$,
\begin{equation}
\label{eq:mi-token}
\E_{z\sim \alpha(\cdot\mid h_t)}
\!\left[\KLop(q_t^z \,\|\, p_t)\right]
=
\KLop(q_t \,\|\, p_t) + I_q(Y_t; Z \mid h_t),
\end{equation}
where $I_q(\cdot;\cdot\mid\cdot)$ is conditional mutual information under the
teacher mixture. The mutual-information term quantifies branch-specific
variation hidden by the marginal teacher target; because it is independent of
$p_t$, the identity is not by itself a different gradient objective. The
corresponding sequence-level identity accumulates this conditional mutual
information across time; Appendix \ref{app:mi-seq} gives the details and
interpretation.

The branch-viability study supplies the task-specific direction: early
high-ambiguity positions are most often labeled real-uncertain, so \pwopsd{}
discounts them with a lower outer weight.

\subsection{PW-OPSD: Position-Weighted On-Policy Self-Distillation}
\label{sec:method:position}

\pwopsd{} uses a deterministic structural proxy for the reliability posterior
$\rho^*(c_t)$: a token's relative position in its own student rollout. For
sequence $i$ with valid completion length $L_i$, define the one-based valid
token index
$t\in\{1,\ldots,L_i\}$ and the per-row position fraction
\[
r_{i,t} = \frac{t-0.5}{L_i}.
\]
The position weight is
\begin{equation}
\label{eq:position-weight}
w_{i,t}
=
w_{\min} + (1-w_{\min})
\sigma\!\left(\frac{r_{i,t}-\tau}{s}\right),
\end{equation}
with defaults $(w_{\min},\tau,s)=(0.25,0.30,0.10)$. The floor keeps early
tokens partially supervised, the threshold places the transition early in the
rollout, and the scale avoids a hard discontinuity. The schedule is increasing
because raw position is positively associated with reliability in the
branch-viability diagnostic.

The per-row normalization in $r_{i,t}$ and the per-sequence reduction are a
single design choice. A position fraction is meaningful only relative to the
sequence's own valid length, not the batch-padded maximum. The corresponding
loss therefore averages within each sequence before averaging across valid
sequences; otherwise, longer rollouts would receive more gradient mass purely
because they contain more tokens.

With $B$ valid sequences in the batch, the implemented \pwopsd{} objective is
\begin{equation}
\label{eq:position-loss}
\mathcal{L}_{\pwopsd{}}
=
\frac{1}{B}\sum_{i=1}^{B}\frac{1}{L_i}\sum_{t=1}^{L_i}
w_{i,t}
\sum_{j\in\V}
\min\!\left(
\opsdtarget_{i,t}(j)
\log\frac{\opsdtarget_{i,t}(j)}
{\student(j\mid x^i_{\mathrm{stu}}, y^i_{<t})},
\tau_{\mathrm{clip}}
\right).
\end{equation}
The inner term is the same per-vocabulary clipped forward KL as
\cref{eq:opsd}. \pwopsd{} uses the ordinary single-pass teacher target
$\opsdtarget_{i,t}$; its additional computation over \opsd{} is only the
scalar sigmoid in \cref{eq:position-weight}.

\section{Experiments}
\label{sec:experiments}

\Cref{sec:method:motivation} suggests a simple reliability prior: early
high-ambiguity branch points are more likely to be teacher-unreliable, while
later positions are more reliable under the branch-viability diagnostic. This
section tests whether the corresponding position-weighted objective improves
downstream reasoning behavior relative to \opsd{} under the same rollout,
teacher pass, and evaluation protocol.

\subsection{Setup}
\label{sec:exp:setup}

\paragraph{Models, baselines, and schedule.}
We train on the Qwen3-4B checkpoint~\citep{qwen3,qwen3_4b_hf} for the
main comparison (\cref{tab:main-4b}), the position-schedule sweep
(\cref{tab:pwopsd-ablation}), and the reduction-positioning ablation
(\cref{tab:reduction-ablation}); the cross-model evidence in
\cref{tab:cross-model} additionally uses DeepSeek-R1-Distill-Llama-8B
(DSR1-L8B)~\citep{deepseekr1,deepseekr1distillllama8b} and
Olmo-3-7B-Think~\citep{olmo37bthink}.
\pwopsd{} uses the Moderate schedule
$(w_{\min},\tau,s)=(0.25,0.30,0.10)$, the diagnostic-derived default. We
compare against three baselines: \opsd{}~\citep{zhao2026opsd} as the
uniform-weight reference, \eopd{}~\citep{jin2026eopd} as a representative
entropy-conditioned adaptive-KL baseline, and
REOPOLD~\citep{ko2026scalingreasoningefficientlyrelaxed} as a cross-family
policy-gradient adaptive on-policy distillation baseline that controls
for whether any adaptive per-token signal recovers the downstream pattern
attributed to position. All methods share the OPSD privileged on-policy
chassis with LoRA~\citep{hu2022lora} (rank $64$, $\alpha=128$), are
evaluated at the $100$-step checkpoint following \opsd{}, and otherwise
use their published defaults. Full training hyperparameters, the
cross-family rationale, and the \pwopsd{} pseudocode are in
\cref{app:evaluation-setup,app:algorithm}.

\paragraph{Evaluation.}
We follow the \opsd{} evaluation setting~\citep{zhao2026opsd} with maximum generation length $38{,}912$ tokens:
\texttt{max\_new\_tokens}$=38912$, \texttt{val\_n}$=12$ samples per problem,
temperature $T=1.0$, top-$p=0.95$, top-$k$ disabled, and
\texttt{enable\_thinking=True}. Benchmarks are
MATH-500~\citep{hendrycks2021math,lightman2024prm800k,math500hf},
AIME 2024~\citep{aime2024hf}, AIME 2025~\citep{aime2025hf}, and HMMT 2025~\citep{dekoninck2026matharena}. We report
$\mathrm{Pass}@12$, $\mathrm{Avg}@12$, and $\mathrm{Maj}@12$;
\cref{app:metrics} gives the formulas. For each method--benchmark
pair we run
three random evaluation seeds (main, $1$, $2$) and report mean $\pm$
across-seed sample standard deviation; the cross-model assessment
(\cref{tab:cross-model}) applies the same four-benchmark three-seed
protocol to two additional checkpoints from different model families.

\subsection{Main results on Qwen3-4B}
\label{sec:exp:scaling}
\begin{table}[t]
\centering
\small
\setlength{\tabcolsep}{4.5pt}
\caption{Qwen3-4B results with maximum generation length $38{,}912$ tokens. Entries report
mean $\pm$ standard deviation across three evaluation seeds. Bold marks the
best value for each benchmark--metric column and any value within $0.1$ pp of
the leader. The Aggressive row is included to show schedule sensitivity rather
than as a separate method claim.}
\label{tab:main-4b}
\begin{tabular}{llccc}
\toprule
Method & Benchmark & Pass@12 & Avg@12 & Maj@12 \\
\midrule
\multirow{4}{*}{\opsd{}~\citep{zhao2026opsd}}
& MATH-500 & $98.00 \pm 0.35$ & $95.33 \pm 0.08$ & $96.73 \pm 0.12$ \\
& AIME 2024 & $86.67 \pm 0.00$ & $75.19 \pm 0.42$ & $\mathbf{80.00 \pm 0.00}$ \\
& AIME 2025 & $\mathbf{83.33 \pm 0.00}$ & $66.67 \pm 1.27$ & $73.33 \pm 3.33$ \\
& HMMT 2025 & $\mathbf{64.44 \pm 1.92}$ & $43.89 \pm 0.73$ & $51.11 \pm 3.85$ \\
\midrule
\multirow{4}{*}{\eopd{}~\citep{jin2026eopd}}
& MATH-500 & $\mathbf{98.33 \pm 0.23}$ & $95.33 \pm 0.08$ & $96.73 \pm 0.23$ \\
& AIME 2024 & $85.56 \pm 1.92$ & $73.61 \pm 2.17$ & $\mathbf{80.00 \pm 0.00}$ \\
& AIME 2025 & $82.22 \pm 1.92$ & $65.65 \pm 0.89$ & $\mathbf{76.67 \pm 3.33}$ \\
& HMMT 2025 & $63.33 \pm 3.33$ & $41.94 \pm 2.00$ & $50.00 \pm 3.33$ \\
\midrule
\multirow{4}{*}{REOPOLD~\citep{ko2026scalingreasoningefficientlyrelaxed}}
& MATH-500 & $\mathbf{98.40 \pm 0.20}$ & $95.09 \pm 0.09$ & $96.53 \pm 0.31$ \\
& AIME 2024 & $86.67 \pm 0.00$ & $73.98 \pm 0.42$ & $\mathbf{80.00 \pm 3.33}$ \\
& AIME 2025 & $81.11 \pm 1.92$ & $62.13 \pm 1.67$ & $74.44 \pm 5.09$ \\
& HMMT 2025 & $58.89 \pm 1.92$ & $41.39 \pm 1.44$ & $51.11 \pm 3.85$ \\
\midrule
\multirow{4}{*}{\pwopsd{} Moderate (ours)}
& MATH-500 & $\mathbf{98.40 \pm 0.00}$ & $95.34 \pm 0.10$ & $96.67 \pm 0.12$ \\
& AIME 2024 & $85.56 \pm 1.92$ & $\mathbf{76.20 \pm 0.58}$ & $\mathbf{80.00 \pm 0.00}$ \\
& AIME 2025 & $\mathbf{83.33 \pm 0.00}$ & $\mathbf{67.78 \pm 1.27}$ & $74.44 \pm 1.92$ \\
& HMMT 2025 & $60.00 \pm 3.33$ & $43.33 \pm 1.21$ & $\mathbf{52.22 \pm 1.92}$ \\
\midrule
\multirow{4}{*}{\pwopsd{} Aggressive (ours)}
& MATH-500 & $\mathbf{98.40 \pm 0.20}$ & $\mathbf{95.53 \pm 0.04}$ & $\mathbf{97.07 \pm 0.12}$ \\
& AIME 2024 & $\mathbf{87.78 \pm 1.92}$ & $75.19 \pm 0.85$ & $\mathbf{80.00 \pm 0.00}$ \\
& AIME 2025 & $\mathbf{83.33 \pm 0.00}$ & $67.59 \pm 0.85$ & $75.56 \pm 1.92$ \\
& HMMT 2025 & $60.00 \pm 0.00$ & $\mathbf{45.37 \pm 0.58}$ & $\mathbf{52.22 \pm 1.92}$ \\
\bottomrule
\end{tabular}
\vspace{-0.3cm}
\end{table}

\Cref{tab:main-4b} reports the Qwen3-4B comparison with maximum generation length $38{,}912$ tokens. We report the diagnostic-derived \emph{Moderate} schedule
$(w_{\min},\tau,s)=(0.25,0.30,0.10)$ as the default \pwopsd{}
configuration. We also include \emph{Aggressive} $(0.05,0.50,0.05)$ from
the schedule sweep as a sensitivity variant that applies a stronger early
discount. The main pattern is that position weighting improves per-sample reasoning accuracy on the AIME benchmarks while preserving near-saturated MATH-500 performance. On MATH-500, \opsd{}, \eopd{}, and \pwopsd{} Moderate are tied within $0.01$ pp Avg@12, while \pwopsd{} Aggressive gives the highest
MATH-500 Avg@12 and Maj@12 in the table. On AIME 2024 and AIME 2025,
\pwopsd{} Moderate improves Avg@12 over \opsd{} by $+1.0$ pp and $+1.1$ pp,
respectively. The local or adaptive alternatives do not reproduce this
Avg@12 pattern: \eopd{} trails \opsd{} by $1.6$ pp on AIME 2024 and $1.0$ pp
on AIME 2025, while REOPOLD trails by $1.2$ pp and $4.5$ pp. HMMT 2025 highlights the role of reliability-curve strength. The Moderate
schedule improves Maj@12 over \opsd{} by $+1.11$ pp but trails by $0.6$ pp
Avg@12 and $4.4$ pp Pass@12. Under the stronger early discount of the
Aggressive schedule, HMMT 2025 Avg@12 rises to $45.37$ pp, a $+1.48$ pp gain
over \opsd{}, and Maj@12 remains $+1.11$ pp above \opsd{}. Among the four evaluated methods we compare, \pwopsd{} is the only one to
improve over \opsd{}: Moderate gains $+1.0$ pp on AIME 2024 and $+1.1$ pp on
AIME 2025, and Aggressive gains $+1.5$ pp on HMMT 2025. \eopd{} and REOPOLD
instead regress on three of four benchmarks, losing $1.1$ and $2.1$ pp Avg@12 on average.

\subsection{Position-schedule ablation}
\label{sec:exp:hyperparam}

\begin{figure}[t]
\noindent
\begin{minipage}[c]{0.36\linewidth}
\raggedright
\includegraphics[width=\linewidth]{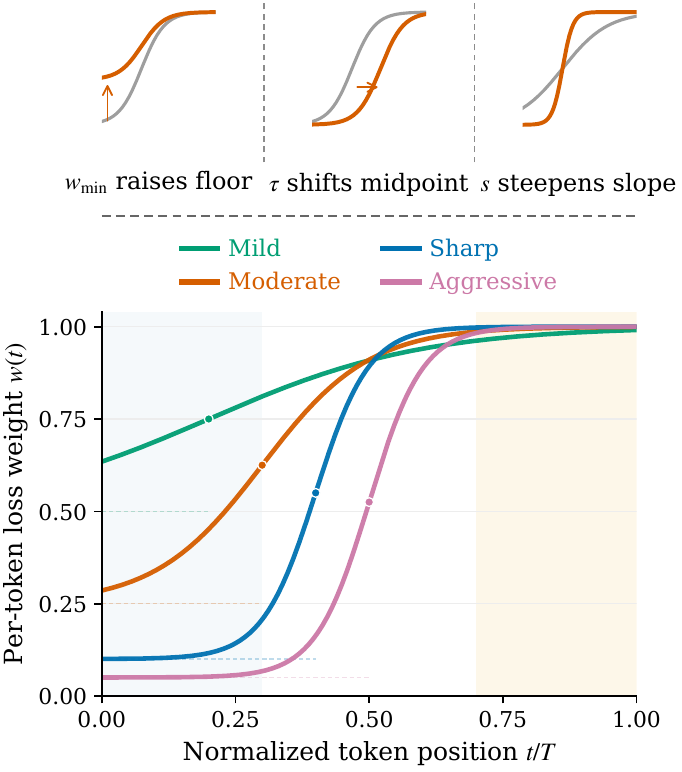}
\end{minipage}%
\hspace{0.02\linewidth}%
\begin{minipage}[c]{0.62\linewidth}
\centering
\scriptsize
\setlength{\tabcolsep}{2.4pt}
\renewcommand{\arraystretch}{0.86}
\begin{tabular}{llccc}
\toprule
Config & Benchmark & Pass@12 & Avg@12 & Maj@12 \\
\midrule
\multirow{4}{*}{Mild}
& MATH-500  & $98.33 \pm 0.12$ & $95.44 \pm 0.05$ & $96.87 \pm 0.12$ \\
& AIME 2024 & $87.78 \pm 3.85$ & $73.98 \pm 0.85$ & $\mathbf{80.00 \pm 0.00}$ \\
& AIME 2025 & $\mathbf{84.44 \pm 1.92}$ & $67.59 \pm 1.16$ & $73.33 \pm 0.00$ \\
& HMMT 2025 & $62.22 \pm 1.92$ & $45.28 \pm 0.73$ & $47.78 \pm 1.92$ \\
\cmidrule(l){2-5}
\multirow{4}{*}{Moderate}
& MATH-500  & $\mathbf{98.40 \pm 0.00}$ & $95.34 \pm 0.10$ & $96.67 \pm 0.12$ \\
& AIME 2024 & $85.56 \pm 1.92$ & $\mathbf{76.20 \pm 0.58}$ & $\mathbf{80.00 \pm 0.00}$ \\
& AIME 2025 & $83.33 \pm 0.00$ & $67.78 \pm 1.27$ & $74.44 \pm 1.92$ \\
& HMMT 2025 & $60.00 \pm 3.33$ & $43.33 \pm 1.21$ & $\mathbf{52.22 \pm 1.92}$ \\
\cmidrule(l){2-5}
\multirow{4}{*}{Sharp}
& MATH-500  & $98.13 \pm 0.31$ & $95.36 \pm 0.15$ & $96.60 \pm 0.20$ \\
& AIME 2024 & $\mathbf{88.89 \pm 7.70}$ & $75.19 \pm 1.53$ & $\mathbf{80.00 \pm 0.00}$ \\
& AIME 2025 & $83.33 \pm 0.00$ & $\mathbf{68.80 \pm 0.89}$ & $\mathbf{75.56 \pm 3.85}$ \\
& HMMT 2025 & $\mathbf{64.44 \pm 1.92}$ & $44.26 \pm 1.12$ & $46.67 \pm 3.33$ \\
\cmidrule(l){2-5}
\multirow{4}{*}{Aggressive}
& MATH-500  & $\mathbf{98.40 \pm 0.20}$ & $\mathbf{95.53 \pm 0.04}$ & $\mathbf{97.07 \pm 0.12}$ \\
& AIME 2024 & $87.78 \pm 1.92$ & $75.19 \pm 0.85$ & $\mathbf{80.00 \pm 0.00}$ \\
& AIME 2025 & $83.33 \pm 0.00$ & $67.59 \pm 0.85$ & $\mathbf{75.56 \pm 1.92}$ \\
& HMMT 2025 & $60.00 \pm 0.00$ & $\mathbf{45.37 \pm 0.58}$ & $\mathbf{52.22 \pm 1.92}$ \\
\bottomrule
\end{tabular}
\end{minipage}

\vspace{6pt}
\refstepcounter{figure}\label{fig:pwopsd-schedules}%
\refstepcounter{table}\label{tab:pwopsd-ablation}%
{\small \textbf{Figure \arabic{figure}:} Per-token weight schedules for the
four configurations; top panels label the role of each $(w_{\min},\tau,s)$
knob. \textbf{Table \arabic{table}:} Position-schedule sweep on Qwen3-4B
(maximum generation length $38{,}912$ tokens). Each cell is mean $\pm$ standard deviation across three
evaluation seeds. Bold marks the best value for each benchmark--metric
column within the sweep. Configurations: Mild $(0.50,0.20,0.20)$,
Moderate $(0.25,0.30,0.10)$, Sharp $(0.10,0.40,0.05)$,
Aggressive $(0.05,0.50,0.05)$.\par}
\vspace{-0.3cm}
\end{figure}

The main comparison fixes the \pwopsd{} schedule to
$(w_{\min},\tau,s)=(0.25,0.30,0.10)$. Since the method introduces three scalar
parameters, we also run a full schedule sweep over four settings to test whether the effect is tied to a single parameter tuple. \Cref{fig:pwopsd-schedules} shows the schedule shapes; \Cref{tab:pwopsd-ablation} reports the completed Qwen3-4B ablation evaluations across all four configurations under the same maximum generation length as the main table.

The four settings vary how strongly early tokens are discounted. Mild
$(0.50,0.20,0.20)$ keeps a high early floor and uses the softest transition;
Moderate $(0.25,0.30,0.10)$ is the chosen configuration from
\Cref{tab:main-4b}; Sharp $(0.10,0.40,0.05)$ lowers the floor and makes the
transition steeper; Aggressive $(0.05,0.50,0.05)$ applies the strongest early
down-weighting and delays the transition furthest. Thus lower $w_{\min}$ makes
the early-token discount more aggressive, larger $\tau$ moves the transition
later, and smaller $s$ makes it sharper.

On MATH-500, every configuration sits within $0.2$ pp Avg@12, well inside across-seed noise. On AIME 2024 Moderate holds the Avg@12 lead but the other three trail by at most $2.2$ pp and all four tie on Maj@12; AIME 2025 Avg@12 is similarly clustered ($67.59$--$68.80$). HMMT 2025 separates the configurations more clearly: Aggressive and Mild gain $+1.9$--$2.0$ pp Avg@12 over Moderate. No configuration dominates uniformly. We retain Moderate in the main table because it is the schedule chosen \emph{a priori} from the diagnostic curves, not because it is the best across this sweep. Averaged across the four benchmarks, all four schedules improve Avg@12 over \opsd{} ($70.27$) by $+0.30$ to $+0.65$ pp and span only $0.35$ pp among themselves, confirming \pwopsd{} is robust to the choice of $(w_{\min},\tau,s)$.

\paragraph{Reduction vs.\ positioning.}
\pwopsd{} differs from \opsd{} along two axes: a position-dependent token
weight $w_t$ (positioning) and a per-rollout average of the token loss before
the batch mean (per-sequence reduction). The full $2{\times}2$ factorial in
\Cref{tab:reduction-ablation} (Appendix~\ref{app:reduction-ablation}) shows
the joint configuration is the only one that matches the AIME 2024 lead of
\Cref{tab:main-4b} ($+1.0$ pp Avg@12 over \opsd{}).

\subsection{Cross-model transfer of the position schedule}
\label{sec:exp:cross-model}
\vspace{-0.1cm}
To test whether the diagnostic-derived Moderate schedule transfers beyond
Qwen3-4B, we evaluate \pwopsd{} on two larger models from different
families: DeepSeek-R1-Distill-Llama-8B (8B, Llama
family)~\citep{deepseekr1,deepseekr1distillllama8b} and Olmo-3-7B-Think
(7B, OLMo family)~\citep{olmo37bthink}. For each model we compare
\pwopsd{} (Moderate, $(w_{\min},\tau,s)=(0.25,0.30,0.10)$) against the
\opsd{} baseline under the same training and evaluation protocols as the
Qwen3-4B main experiments: the full four-benchmark suite
(MATH-500, AIME 2024, AIME 2025, HMMT 2025), three independent evaluation
seeds per cell, and the maximum generation length of $38{,}912$ tokens described in
\cref{sec:exp:setup}. The Moderate schedule is held fixed across all three
models with no per-model retuning. For DSR1-L8B and Olmo-3-7B-Think, both
trained as reasoning models, we additionally apply a tokenizer-level
$\langle\text{think}\rangle$-closure that replicates
\texttt{enable\_thinking=False} on chat templates that ignore the flag,
matching the training-time student prompt format used on Qwen3-4B
(evaluation continues to use \texttt{enable\_thinking=True} as described
in \cref{sec:exp:setup}).

\begin{table}[t]
\centering
\small
\setlength{\tabcolsep}{3.5pt}
\renewcommand{\arraystretch}{1.05}
\caption{Cross-model Avg@12 evidence for \pwopsd{} (Moderate, no per-model
retuning) versus the \opsd{} baseline of \citet{zhao2026opsd} across three
model families with maximum generation length $38{,}912$ tokens. Entries report
mean $\pm$ across-seed sample standard deviation over three evaluation
seeds; the Avg@12 column is the equal-weight mean across the four
benchmarks, computed from unrounded seed-level values. Bold marks the within-model winner in each numeric column,
including values within $0.1$ pp of the leader.}
\label{tab:cross-model}
\resizebox{\linewidth}{!}{%
\begin{tabular}{llccccc}
\toprule
Model & Method & MATH-500 & AIME 2024 & AIME 2025 & HMMT 2025 & Avg@12 \\
\midrule
\multirow{2}{*}{Qwen3-4B}
  & \opsd{}
    & $\mathbf{95.33 \pm 0.08}$
    & $75.19 \pm 0.42$
    & $66.67 \pm 1.27$
    & $\mathbf{43.89 \pm 0.73}$
    & $70.27$ \\
  & \pwopsd{} (\textbf{ours})
    & $\mathbf{95.34 \pm 0.10}$
    & $\mathbf{76.20 \pm 0.58}$
    & $\mathbf{67.78 \pm 1.27}$
    & $43.33 \pm 1.21$
    & $\mathbf{70.66}$ \\
\cmidrule(l){2-7}
\multirow{2}{*}{DSR1-L8B}
  & \opsd{}
    & $\mathbf{89.02 \pm 0.19}$
    & $40.74 \pm 2.58$
    & $31.57 \pm 1.28$
    & $20.93 \pm 0.80$
    & $45.56$ \\
  & \pwopsd{} (\textbf{ours})
    & $88.72 \pm 0.25$
    & $\mathbf{41.20 \pm 0.32}$
    & $\mathbf{32.22 \pm 0.56}$
    & $\mathbf{21.48 \pm 0.42}$
    & $\mathbf{45.91}$ \\
\cmidrule(l){2-7}
\multirow{2}{*}{Olmo-3-7B-Think}
  & \opsd{}
    & $\mathbf{96.20 \pm 0.08}$
    & $\mathbf{72.78 \pm 0.28}$
    & $67.41 \pm 2.16$
    & $\mathbf{43.89 \pm 0.73}$
    & $70.07$ \\
  & \pwopsd{} (\textbf{ours})
    & $\mathbf{96.26 \pm 0.19}$
    & $72.59 \pm 1.53$
    & $\mathbf{69.63 \pm 1.12}$
    & $\mathbf{43.80 \pm 0.58}$
    & $\mathbf{70.57}$ \\
\bottomrule
\end{tabular}%
}
\vspace{-0.3cm}
\end{table}

\Cref{tab:cross-model} shows that the same Moderate schedule gives positive
Avg@12 gains on every tested model: $+0.39$ pp on Qwen3-4B (4B parameters,
Qwen family), $+0.35$ pp on DeepSeek-R1-Distill-Llama-8B (8B, Llama family),
and $+0.50$ pp on Olmo-3-7B-Think (7B, OLMo family). The pattern spans
Qwen, Llama, and OLMo checkpoints across a 4B--8B parameter range, with no
per-model schedule retuning, supporting the model-family portability of the
position-rank reliability signal identified in
\cref{sec:method:motivation}.

\vspace{-0.3cm}
\section{Conclusion, Limitations, and Future Work}
\label{sec:conclusion}
\vspace{-0.3cm}
In this paper, we study teacher-token reliability in on-policy
self-distillation. To address whether high teacher entropy
reflects low reliability or benign solution diversity, we design a
branch-viability diagnostic, showing empirically that reliability is positionally
structured (AUROC $0.83$ vs.\ $\le 0.57$ for local uncertainty). Inspired by
this finding, we introduce \pwopsd{}, an increasing-sigmoid position weight
on the \opsd{} chassis. Our experiments show that \pwopsd{} improves Avg@12
over \opsd{} on AIME 2024 ($+1.0$ pp), AIME 2025 ($+1.1$ pp), and on
DeepSeek-R1-Distill-Llama-8B and Olmo-3-7B-Think. One limitation is that \pwopsd{} is
plug-and-play yet its gains over \opsd{} remain modest, leaving room to
design more sophisticated methods built on the positional reliability
finding. To improve the proposed method, we aim to design more dedicated
objectives that go beyond pure position weighting, such as
position-conditioned mixing of forward and reverse KL.

\bibliographystyle{plainnat}
\bibliography{references}

\newpage
\appendix
\crefalias{section}{appendix}
\crefalias{subsection}{appendix}
\section{Branch-viability protocol details}
\label{app:branch-viability}

\paragraph{Models, prompts, and software.}
The diagnostic uses one \texttt{Qwen/Qwen3-4B}
checkpoint~\citep{qwen3_4b_hf} (HuggingFace snapshot
\texttt{1cfa9a72\dots3b3df60c}, dtype \texttt{bfloat16}) with no adapters
loaded. ``Teacher'' and ``student'' denote two prompt templates applied to
this single checkpoint: the teacher template includes the privileged
ground-truth answer, and the student template is the ordinary problem
prompt. vLLM and HuggingFace load this checkpoint as separate software
backends, not as different trained weights. Problems are drawn from three sources, sampled without replacement (see
attrition below): \texttt{MATH-500} test
split~\citep{hendrycks2021math,lightman2024prm800k,math500hf},
\texttt{AIME 2024}~\citep{aime2024hf}, and
\texttt{AIME 2025}~\citep{aime2025hf}.
We use \texttt{vllm 0.11.0} with \texttt{tensor\_parallel\_size=4} (TP$=4$)
for generation~\citep{kwon2023vllm}, \texttt{transformers 4.57.1} for the
HuggingFace (HF) teacher forward passes~\citep{wolf2020transformers}, and
\texttt{torch 2.8.0+cu128} on $4{\times}$H100 80GB
GPUs~\citep{paszke2019pytorch}.

\paragraph{Problem attrition.}
Phase A samples $24$ MATH-500, $30$ AIME 2024, and $30$ AIME 2025 problems
($84$ total). Phase B produces $23 + 21 + 18 = 62$ correct-spine problems.
After Phase F labeling, one AIME 2024 correct-spine problem has only gray
candidates, so the binary-labeled pool used in \cref{tab:branch-auroc}
contains $23 + 20 + 18 = 61$ usable problems. Phase C then proposes up to $5$
high-truncated-entropy candidates per problem (with the spacing /
plausibility filters below); Phase E rolls out $6$ student continuations
per (candidate, alternative) pair. The \cref{tab:branch-auroc} report
restricts to high-truncated-entropy candidate positions and to the $8$
candidates labeled \emph{real-uncertain} plus $271$ \emph{diversity}
candidates across these $61$ problems.
\Cref{fig:branch-auroc} uses the broader continuous-viability subset before
binary thresholding.

\paragraph{Random seeds.}
Phase A fallback sampling, Phase E forced continuations, and the
$\widetilde r$ residualization use vLLM seed \texttt{0}; Phase D MC-dropout
masks use a per-position random state derived from the model's default
generator. The cluster bootstrap in Phase G uses NumPy seed \texttt{0}.

\paragraph{Score residualization and AUROC.}
Within each problem $p$, every raw uncertainty score $u$ (or its
sign-oriented form for position) is mean-centered: $u' = u - \bar{u}_p$.
We compute AUROC on the residualized scores against the binary label
defined in Phase F, using the standard $\mathrm{Mann\text{-}Whitney}\,U$
formula with mid-rank tie handling. The cluster bootstrap resamples
problems (not candidates) with replacement and recomputes the
within-problem residualization on each resample, giving the $95\%$
intervals reported in \cref{tab:branch-auroc}.

\begin{itemize}\itemsep=0pt
\item \textbf{Phase A.} vLLM (TP=$4$) student greedy decode of $84$ problems
in total ($24$ from MATH-500, $30$ from AIME 2024, and $30$ from AIME 2025;
the protocol is run independently per dataset and the labeled candidates
are pooled in Phase G), with fallback to $T{=}0.7$ for problems where
greedy does not reach $\backslash\mathrm{boxed}$ within $16$K tokens.
\item \textbf{Phase B.} HF forward of the same checkpoint under the teacher
template (privileged-answer prompt $+$ spine), giving per-position teacher
distributions and top-$M{=}16$ valid-token truncated entropy.
\item \textbf{Phase C.} Apply junk / EOS / special filters, plausibility
filter $p_2 \ge 0.02 \wedge p_2/p_1 \ge 0.10$, $64$-token spacing
constraint, and skip post-$\backslash\mathrm{boxed}$ positions. Keep up to
$5$ high-truncated-entropy candidate positions per problem; for each, record
the teacher pass's top-$3$ valid child tokens as forced alternatives.
\item \textbf{Phase D.} HF teacher MC dropout ($M{=}5$, $p{=}0.1$, last $4$
layers) at each candidate; record $H_{\mathrm{full}}$, MI, and
$\log\hat{\kappa}$.
\item \textbf{Phase E.} vLLM (TP=$4$) \emph{student}-context forced
continuation: each forced child token from Phase C is appended to the
student-template prompt $+$ spine truncated to the candidate position, and
rollouts continue under the student template (no privileged information).
We use student-template continuation rather than teacher-template
continuation because a teacher-template continuation can re-use the
privileged answer in its prompt and recover from almost any forced child,
collapsing the labels toward diversity. Across the three datasets, up to
$(24+30+30) \times 5 \times 3 \times 6 = 7560$ attempted rollouts at
$T{=}1.0$, \texttt{top\_p}$=0.95$, with \texttt{max\_tokens} dynamically
clipped per request to fit \texttt{max\_model\_len}$=32$K.
\item \textbf{Phase F.} For each forced child, define viability as the
fraction of its $6$ student-context continuations whose extracted boxed
answer matches the ground truth. Label a candidate \emph{diversity} if at
least two of its children have viability $\ge V_{\mathrm{high}}{=}0.75$;
label it \emph{real-uncertain} if every child has viability $< V_{\mathrm{low}}{=}0.40$
and the mean child viability is $< V_{\mathrm{low}}$; otherwise label
``gray'' and exclude from the AUROC.
\item \textbf{Phase G.} AUROC + area-under-precision-recall-curve (AUPRC) + cluster-bootstrap by problem
($2000$ resamples, multiplicity-preserved); scatter and histograms.
\end{itemize}

\paragraph{Privileged-info teacher prompt.}
We use the following template (wrapped in the Qwen3 chat template with
\texttt{enable\_thinking=True}):
\begin{quote}\small\itshape
You are a privileged teacher. Solve the problem and end your solution with
the correct boxed answer.\\
Problem: \{problem\}\\
Privileged ground-truth final answer: \{answer\}\\
Use the ground-truth answer above. Produce a complete step-by-step solution
that ends with $\backslash\mathrm{boxed}\{\{$answer$\}\}$.
\end{quote}
This is a simplified privileged-info injection in the spirit of \opsd{}'s
training-time teacher prompt; it is not pixel-identical to the \opsd{}
official template but conditions the teacher's autoregressive distribution on
the ground-truth answer for the duration of generation.

\section{MC-dropout diagnostic implementation}
\label{app:mc-dropout}

MC dropout is diagnostic-only in this paper. It is used to compute
$H_{\mathrm{full}}$, MI, and $\log\hat{\kappa}$ scores for
\cref{tab:branch-auroc}; it is not used to construct a training target or a
token weight. All training methods use the ordinary single-pass teacher
target $\opsdtarget_t$.

\paragraph{MC dropout via forward hooks.}
Modern Qwen / Llama transformer blocks do not expose an \texttt{nn.Dropout}
submodule. We inject MC dropout at inference time via forward hooks on the
last $L=4$ transformer layers; the hook applies
\(\mathrm{F.dropout}(h, p, \mathrm{training}{=}\mathrm{True})\)
to the layer output. We use $p=0.1$ throughout and $M=5$ MC samples. The
unperturbed privileged-teacher forward provides $\opsdtarget_t$; the
perturbed forwards are used only for the diagnostic uncertainty scores. Hooks
are removed in a \texttt{try/finally} block to prevent leaking dropout into
subsequent forward passes.

\paragraph{Numerical stability of $\hat{\kappa}_t$.}
The moment-matching estimator for $\hat{\kappa}_t$ can produce negative
values when the sample variance trace exceeds the categorical variance trace
$1 - \|\bar{\mathbf p}_t\|_2^2$ (an artifact of small $M$). For
\cref{tab:branch-auroc}, we do not apply any training-style floor or
log-normalization. We store the raw $\hat{\kappa}_t$ estimate and compute
$\log \max(\hat{\kappa}_t,\epsilon)$ only at score time, with
$\epsilon=10^{-6}$ for the log transform.

\paragraph{Tail-bucket aggregation.}
For very large vocabularies (Qwen3 vocab $\approx 152$K), computing the full
softmax per MC sample at every position is memory-bound. We compute moment
statistics token-by-token and aggregate
$\sum_m \|\mathbf{p}^{(m)}_t\|_2^2$ incrementally to avoid materializing the
full per-sample distribution beyond one forward at a time. In Phase B of the
branch-viability experiment we still materialize the per-position softmax
once for top-$M$ selection; this is a one-time pass per problem.

\section{Reliability-weighted surrogate details}
\label{app:reliability-surrogate}

\Cref{sec:method:theory} conditions the reliability posterior on the
distillation context $c_t=(x_{\mathrm{stu}},x_{\mathrm{tch}},y_{<t})$. This
conditioning makes the tower-property step explicit. For fixed $\theta$ and a
fixed teacher, the token divergence
$D_t(\theta)=\KLop(q_t\,\|\,p_t)$ is determined by $c_t$, so
\[
\E[I_tD_t(\theta)]
=
\E\!\left[\E[I_tD_t(\theta)\mid c_t]\right]
=
\E\!\left[D_t(\theta)\Pr(I_t=1\mid c_t)\right].
\]
Conditioning only on the student prefix $h_t=(x_{\mathrm{stu}},y_{<t})$ would
require an additional assumption, because the privileged teacher target can
vary with the teacher prompt and reference information even when the student
prefix is fixed.

The risk in \cref{eq:reliability-risk} is also a distillation surrogate on
sampled rollouts. As in standard \opsd{} implementations, the rollout is
sampled from the current student and then treated as a fixed training example
for the KL update; gradients pass through the student probabilities on visited
prefixes, not through the sampling operation that produced those prefixes. The
main-text interpretation uses the unclipped divergence $D_t(\theta)$ for
notational clarity, while the implemented objectives use the element-wise
clipped forward-KL surrogate described in \cref{eq:opsd}.

\section{Branch-mixture identity and interpretation}
\label{app:mi-seq}

\Cref{eq:mi-token} gives the token-level identity used in the main text. With
joint distribution
$q(z,y_t\mid h_t)=\alpha(z\mid h_t)q_t^z(y_t)$ and marginal
$q_t(y_t)=\sum_z\alpha(z\mid h_t)q_t^z(y_t)$,
\[
\begin{aligned}
\E_{z\sim\alpha}\KLop(q_t^z\,\|\,p_t)
&=
\sum_{z,y_t}q(z,y_t\mid h_t)
\log\frac{q_t^z(y_t)}{p_t(y_t)}\\
&=
\KLop(q_t\,\|\,p_t)
+ I_q(Y_t;Z\mid h_t).
\end{aligned}
\]
The mutual-information term is independent of $p_t$. The identity therefore
decomposes branch-specific variation hidden by the marginal teacher target; it
is not, by itself, a proof that marginal forward KL has a different gradient
objective. The empirical branch-viability diagnostic supplies the task-specific
direction used by \pwopsd{}: early high-ambiguity positions more often
correspond to unreliable supervision.

The same branch-mixture view yields a sequence-level form. Let the teacher's
branch-conditioned next-token distribution be
$q_t^z(y_t \mid y_{<t})$, and let the corresponding sequence distribution
factorise autoregressively as
\[
q^z(y_{1:T}) = \prod_{t=1}^{T} q_t^z(y_t \mid y_{<t}) .
\]
Its marginal under the latent branch prior $\alpha$ is
$q(y_{1:T}) = \sum_z \alpha(z)\, q^z(y_{1:T})$. For a student sequence
distribution
$p(y_{1:T}) = \prod_{t=1}^{T} p_t(y_t \mid y_{<t})$,
\begin{equation}
\label{eq:mi-seq}
\E_{z\sim \alpha}
\!\left[\KLop(q^z \,\|\, p)\right]
=
\KLop(q \,\|\, p) + \sum_t I_q(Y_t; Z \mid Y_{<t}).
\end{equation}
Thus the value of branch-conditioned sequence forward KL differs from marginal
sequence forward KL by the sum of conditional mutual-information terms between
the next token and the latent branch. As in the token-level identity, this sum
quantifies branch ambiguity but is independent of the student distribution.

\section{Position-schedule hyperparameters}
\label{app:position-sweep}

The Moderate schedule used in the main table sets
$(w_{\min},\tau,s)=(0.25,0.30,0.10)$, placing the transition near the
empirical early-to-late change in the branch-viability diagnostic while
keeping a nonzero floor on early-token supervision. The full position
sweep in \cref{tab:pwopsd-ablation} covers four schedules:
\textbf{Mild} $(0.50,0.20,0.20)$,
\textbf{Moderate} $(0.25,0.30,0.10)$ (the a-priori choice from the
diagnostic curve in \cref{fig:branch-auroc}),
\textbf{Sharp} $(0.10,0.40,0.05)$, and
\textbf{Aggressive} $(0.05,0.50,0.05)$. The main table reports both
Moderate (the a-priori headline) and Aggressive (the sweep configuration
that recovers the HMMT 2025 cell on Avg@12). We did not tune $(w_{\min},
\tau, s)$ on any evaluation benchmark; the four configurations were
chosen before observing any of the Avg@12/Pass@12/Maj@12 values.

\section{Adaptive-loss template and method comparison}
\label{app:method-template}

\Cref{tab:method-template} compares the three per-token
distribution-matching objectives evaluated in \cref{tab:main-4b} along the
dimensions that the template makes explicit: target distribution, adaptive
signal or reliability proxy, and reduction over tokens.
REOPOLD~\citep{ko2026scalingreasoningefficientlyrelaxed} is also
evaluated in \cref{tab:main-4b}; it does not fit this template because
its gradient flows through the rolled-out token's log-prob rather
than through a distributional forward KL. Its role in the comparison is
summarized in \cref{sec:method:adaptive}.

\subsection{Comparison to Related Adaptive Losses}
\label{sec:method:adaptive}

\Cref{alg:position} (\cref{app:algorithm}) gives the explicit one-step
pseudocode. For reference, the scalar-weighted clipped-FKL template is
\begin{equation}
\label{eq:adaptive}
\mathcal{L}_{\mathrm{wFKL}}
=
\frac{1}{|\mathcal{M}|}
\sum_{t\in\mathcal{M}}
w_t
\sum_{j\in\V}
\min\!\left(
\teacher_t(j)\log\frac{\teacher_t(j)}{\student(j\mid x_{\mathrm{stu}},y_{<t})},
\tau_{\mathrm{clip}}
\right).
\end{equation}
Different adaptive losses instantiate or modify this template through the
choice of target, reliability proxy, inner divergence, and reduction.

\opsd{} uses a uniform weight~\citep{zhao2026opsd}. \eopd{}~\citep{jin2026eopd} gates a forward-KL
augmentation to reverse KL using teacher entropy. \pwopsd{} keeps the OPSD
forward-KL inner loss and uses the position schedule $w_{i,t}$ from
\cref{eq:position-weight} in its outer token aggregation. REOPOLD is also
included in the experimental comparison but is a policy-gradient distillation
variant rather than a per-token forward-KL weighting rule~\citep{ko2026scalingreasoningefficientlyrelaxed},
so it does not instantiate the template above; \cref{sec:exp:setup} states the
cross-family rationale. \Cref{tab:method-template}
(Appendix~\ref{app:method-template}) lays out the three per-token
distribution-matching objectives in a side-by-side table.

\begin{table}[h]
\centering
\small
\setlength{\tabcolsep}{4pt}
\caption{Per-token distribution-matching distillation objectives compared in
\cref{tab:main-4b}, parametrized by target distribution, adaptive signal or
reliability proxy, and reduction over tokens. REOPOLD is
also evaluated in \cref{tab:main-4b} but follows a policy-gradient
route rather than a per-token forward-KL weighting, so it is not
instantiated in this template.}
\label{tab:method-template}
\begin{tabular}{lccc}
\toprule
Method & target / inner loss & adaptive signal / proxy & reduction\\
\midrule
\opsd{}~\citep{zhao2026opsd}  & clipped FKL to $\opsdtarget_t$       & uniform ($w_t=1$)  & global token mean\\
\eopd{}~\citep{jin2026eopd}   & entropy-gated RKL/FKL mixture        & teacher-entropy gate & global token mean\\
\pwopsd{} (\textbf{ours})     & clipped FKL to $\opsdtarget_{i,t}$   & position $w_{i,t}$ & per-sequence mean\\
\bottomrule
\end{tabular}
\end{table}

\section{PW-OPSD training pseudocode}
\label{app:algorithm}

\Cref{alg:position} gives one training step of \pwopsd{}. The procedure
computes student and teacher log-probabilities at the distillation
temperature, forms the forward-KL tensor with no reduction, clamps each
vocabulary element, sums over the vocabulary, applies the position weight,
averages over valid tokens within each sequence, and then averages over
valid sequences. The sampled rollout is fixed for this update, as discussed in
\cref{app:reliability-surrogate}.

\begin{algorithm}[h]
\caption{One training step of \pwopsd{}.}
\label{alg:position}
\begin{algorithmic}[1]
\Require Parameters $\theta$, prompts $\{(x^i_{\mathrm{stu}},x^i_{\mathrm{tch}})\}_{i=1}^{B}$, schedule $(w_{\min},\tau,s)$, distillation temperature $T_{\mathrm{distill}}$, clip $\tau_{\mathrm{clip}}$.
\State Sample student rollouts $y^i \sim \student(\cdot\mid x^i_{\mathrm{stu}})$.
\State Let $L_i$ be the number of valid generated tokens before EOS or truncation.
\State Score visited prefixes at $T_{\mathrm{distill}}$ under the privileged teacher context and ordinary student context to obtain $\opsdtarget_{i,t}$ and $p_{i,t}$ for each valid $t$.
\For{each sequence $i$ and valid token $t\in\{1,\ldots,L_i\}$}
    \State $r_{i,t}\gets(t-0.5)/L_i$
    \State $w_{i,t}\gets w_{\min}+(1-w_{\min})\sigma((r_{i,t}-\tau)/s)$
    \State $\ell_{i,t}\gets\sum_{j\in\V}\min\!\left(\opsdtarget_{i,t}(j)\log\frac{\opsdtarget_{i,t}(j)}{p_{i,t}(j)},\,\tau_{\mathrm{clip}}\right)$
\EndFor
\State $\mathcal{L}_{\pwopsd{}}\gets B^{-1}\sum_{i=1}^{B}L_i^{-1}\sum_{t=1}^{L_i}w_{i,t}\ell_{i,t}$
\State Update $\theta$ using $\nabla_\theta\mathcal{L}_{\pwopsd{}}$.
\end{algorithmic}
\end{algorithm}

\section{Implementation conventions inherited from OPSD}
\label{app:opsd-conventions}

The following implementation conventions are held fixed across the methods
reported in this paper.

\paragraph{Right-padded prompt collator + batch-max loss slicing.}
The upstream \opsd{} \texttt{data\_collator} right-pads prompts and the
trainer slices loss only on the first \texttt{batch\_max\_prompt\_len}
tokens of each completion; this is held constant across all methods in
\cref{tab:main-4b}. OpenThoughts-Math-30k~\citep{openthoughtsmath30kopsd} prompts vary in length (median
$93$ tokens, max $826$), so right-padding produces a per-batch prompt-PAD
gap. Methods are compared on the same gap.

\paragraph{Train/eval prompt template gap.}
Training prompts use the \opsd{} reference student template
\texttt{Problem: \{problem\}\textbackslash n\textbackslash n Please reason...}
with \texttt{enable\_thinking=False}; evaluation prompts use the simpler
form
\texttt{\{problem\}\textbackslash n\textbackslash n Please reason...}
with \texttt{enable\_thinking=True}. This training/evaluation prompt-template difference is shared
across all methods reported in this paper.

\paragraph{Per-vocabulary clip semantics.}
The clipped forward KL in \cref{eq:opsd} clamps each token/vocabulary entry
of $q_t(j)\log(q_t(j)/p_t(j))$ element-wise via
\texttt{F.kl\_div(reduction='none').clamp(max=tau\_clip)} before the inner
sum over the vocabulary. This matches the \opsd{} reference implementation.

\paragraph{Gradient-accumulation token-mean.}
With per-microbatch token-mean reduction and gradient accumulation $=2$, the
\opsd{} loss computes the mean of two per-microbatch token-means rather than
the exact token-weighted mean over the effective batch. This is the standard
HuggingFace Trainer behavior and is held fixed for the \opsd{} baseline.
\pwopsd{}'s per-sequence reduction does not have this issue because each
microbatch contains the same number of sequences.

\paragraph{vLLM rollout seed.}
The vLLM colocate-mode rollout sampler is seeded as
\texttt{accelerator.process\_index // tp\_size}, independent of the trainer
\texttt{--seed} flag. Reruns of the same \texttt{--seed N} therefore produce
slightly different rollouts. All evaluation results use seeded vLLM
\texttt{SamplingParams.seed}, which is reported per evaluation run.

\section{Evaluation setup}
\label{app:evaluation-setup}
\paragraph{Models.}
The Qwen3-4B checkpoint~\citep{qwen3,qwen3_4b_hf} is used in the main
comparison (\cref{tab:main-4b}), the position-schedule sweep
(\cref{tab:pwopsd-ablation}), and the reduction-positioning ablation
(\cref{tab:reduction-ablation}); the
cross-model evidence in \cref{tab:cross-model} additionally uses
DeepSeek-R1-Distill-Llama-8B~\citep{deepseekr1,deepseekr1distillllama8b}
and Olmo-3-7B-Think~\citep{olmo37bthink}, two larger models from
different families. In every comparison we train LoRA
adapters on a fixed checkpoint and merge the selected adapter before
evaluation, so within each model block, differences between method rows
come only from the distillation objective and not from the checkpoint
weights.

\paragraph{Training.}
We use the \opsd{} privileged on-policy setup~\citep{zhao2026opsd}. Across
methods, the local implementation holds fixed preprocessing and prompt
templates, full-vocabulary clipped forward-KL conventions, LoRA~\citep{hu2022lora}
rank $64$, $\alpha=128$, dropout $0.05$, learning rate $5{\times}10^{-6}$,
\texttt{max\_completion\_length}$=1024$, distillation temperature
$T_{\mathrm{distill}}=1.1$, KL clip $\tau_{\mathrm{clip}}=0.05$, a fixed teacher
with LoRA disabled in the privileged forward pass, and seed $42$. The local
launcher uses $4{\times}$H100 GPUs with effective batch size $32$ (per-device
batch $4$ with gradient accumulation $2$). All methods are evaluated at
the $100$-step checkpoint, following the OPSD evaluation horizon
of~\citet{zhao2026opsd}. \pwopsd{} uses
$(w_{\min},\tau,s)=(0.25,0.30,0.10)$ for the diagnostic-derived default
schedule.

\paragraph{Baselines.}
We compare \pwopsd{} against three baselines run under our common
training and evaluation protocol: \opsd{}~\citep{zhao2026opsd} as the
uniform-weight reference, \eopd{}~\citep{jin2026eopd} as a
representative entropy-conditioned adaptive-KL alternative,
and REOPOLD~\citep{ko2026scalingreasoningefficientlyrelaxed} as a
representative policy-gradient adaptive on-policy distillation method.
REOPOLD differs from the per-token forward-KL family in gradient form,
because its gradient flows through the rolled-out token's log-prob
rather than through a distributional forward KL. It nevertheless answers the
same operational question of how to derive a per-token training signal from
teacher-student log-likelihood-ratio rewards on sampled rollout tokens.
Including REOPOLD therefore provides a cross-family
reference point that controls for the possibility that any adaptive
token weighting recovers the downstream pattern attributed to
position. All baselines use their published default hyperparameters
under the common training and evaluation protocol described above.

\paragraph{Evaluation setting.}
We use a maximum generation length of $38{,}912$ tokens:
\texttt{max\_new\_tokens}$=38912$, $N=12$ samples per problem, temperature
$T=1.0$, top-$p=0.95$, top-$k$ disabled, and
\texttt{enable\_thinking=True}, matching the \opsd{} evaluation
setup~\citep{zhao2026opsd}.

\paragraph{Metrics.}
We report $\mathrm{Avg}@12$, $\mathrm{Pass}@12$, and $\mathrm{Maj}@12$, which
together expose per-sample accuracy, search-style success under repeated
attempts, and stability under aggregation. Majority vote uses
math-equivalence clustering with \texttt{sympy}~\citep{meurer2017sympy} and
\texttt{math\_verify}~\citep{mathverify2025}; unformatted predictions are
placed in a single \texttt{INVALID} cluster that participates in the
plurality count and is scored incorrect when selected. Formal per-problem
definitions are deferred to \cref{app:metrics}.

\paragraph{Benchmarks.}
We evaluate on MATH-500~\citep{hendrycks2021math,lightman2024prm800k,math500hf},
AIME 2024~\citep{aime2024hf}, AIME 2025~\citep{aime2025hf}, and HMMT
February 2025. The HMMT set is a locally cleaned parquet derived from
MathArena's \texttt{hmmt\_feb\_2025} release~\citep{dekoninck2026matharena}, with
SHA-256 recorded in the appendix. For each method--benchmark pair we run three random
evaluation seeds (main, $1$, $2$) and report mean $\pm$ across-seed sample
standard deviation. The cross-model assessment (\cref{tab:cross-model})
extends this protocol to DeepSeek-R1-Distill-Llama-8B and Olmo-3-7B-Think
under the same four benchmarks.

\section{Evaluation metric definitions}
\label{app:metrics}

\paragraph{Multi-sample evaluation for reasoning.}
Reasoning models are commonly evaluated with repeated sampling because a
single completion can understate the chance that the model finds a correct
solution. Pass@N measures whether any of $N$ samples succeeds
\citep{chen2021codex,brown2024scaling}, while self-consistency and majority
vote aggregate multiple reasoning paths~\citep{wang2023selfconsistency}. We
report Avg@12, Pass@12, and Maj@12 throughout. These metrics expose different
behaviors of a distillation objective: per-sample accuracy, search-style
success under repeated attempts, and stability under aggregation.

\paragraph{Formal definitions.}
For a single problem with $N{=}12$ generated solutions
$\{y^{(1)},\ldots,y^{(N)}\}$, the predicted answer is extracted as the
content of the last $\texttt{\textbackslash boxed\{...\}}$ in $y^{(i)}$;
samples without a parseable boxed answer are assigned the cluster key
$\texttt{INVALID}$. Let $c_i \in \{0,1\}$ indicate whether $y^{(i)}$ is
graded correct against the gold answer (using
\texttt{math\_verify}~\citep{mathverify2025} with a normalized
string-equality fallback for parsing failures), and let $a_i$ be the
math-equivalence cluster key obtained by grouping the extracted answers
under the \texttt{sympy}~\citep{meurer2017sympy} and \texttt{math\_verify}
normalization pipeline. The per-problem metrics are
\begin{align}
\mathrm{Avg}@N  &= \frac{1}{N}\sum_{i=1}^{N} c_i,\\
\mathrm{Pass}@N &= \mathbf{1}\!\left[\textstyle\sum_{i=1}^{N} c_i \ge 1\right],\\
\mathrm{Maj}@N  &=
\begin{cases}
0,            & a^\star = \texttt{INVALID},\\
c_{i^\star},  & \text{otherwise},
\end{cases}
\end{align}
where the plurality cluster key is
$a^\star \in \arg\max_{a \in \{a_1,\ldots,a_N\}} \#\{j : a_j = a\}$, with
ties broken by smallest first occurrence
$a^\star = a_{\min\{i : a_i \in \arg\max\}}$, and the representative index
$i^\star = \min\{i : a_i = a^\star\}$. The $\texttt{INVALID}$ cluster
\emph{participates in the plurality count} but is scored incorrect when
selected, so a problem on which the model produces no parseable answer in
the majority of samples scores $\mathrm{Maj}@N=0$ even if some minority of
samples were correct. Reported numbers are means of these per-problem
metrics across the benchmark, optionally averaged across evaluation seeds
with a sample (Bessel-corrected) standard deviation.

\section{Reproducibility notes}
\label{app:reproducibility}

% An anonymized release of the training and evaluation code accompanying
% this submission is available at
% \url{https://anonymous.4open.science/r/pw_opsd_oss-F07B}.
The training and evaluation code accompanying
this paper is available at
\url{https://github.com/SaFo-Lab/PW-OPSD}. The main
dataset sources and evaluation conventions are as follows.

\begin{itemize}\itemsep=0pt
\item \textbf{Datasets.} MATH-500 is HuggingFace
\texttt{HuggingFaceH4/MATH-500}~\citep{hendrycks2021math,lightman2024prm800k,math500hf};
AIME 2024 is HuggingFace
\texttt{HuggingFaceH4/aime\_2024}~\citep{aime2024hf}; AIME 2025 is HuggingFace
\texttt{yentinglin/aime\_2025}~\citep{aime2025hf}; HMMT February 2025 is a locally cleaned
parquet derived from the MathArena \texttt{hmmt\_feb\_2025}
release~\citep{dekoninck2026matharena}, SHA-256
\texttt{87bfb23d2c887fab12b42fcc2b2dd8cb5a9d1070e591490dac8755bc366ea25e}; the training corpus is
OpenThoughts-Math-30k~\citep{openthoughts2025,openthoughtsmath30kopsd}, the same corpus used by
\citet{zhao2026opsd}.
\item \textbf{Maj@N evaluation.} Maj@N clusters all $N$ predictions by
math-equivalence using \texttt{sympy}~\citep{meurer2017sympy} and
\texttt{math\_verify}~\citep{mathverify2025}, with unformatted predictions
placed in a single \texttt{INVALID} cluster that participates in the
plurality count and is scored incorrect when selected
(see \cref{app:metrics} for the formal definition).
\end{itemize}

\section{Reduction $\times$ positioning ablation}
\label{app:reduction-ablation}

\pwopsd{} differs from \opsd{} along two axes that are easy to conflate. The
first is \emph{positioning}: \pwopsd{} multiplies the per-token loss by a
position-dependent scalar $w_t \in [w_{\min}, 1]$, while \opsd{} uses
$w_t \equiv 1$. The second is \emph{reduction}: \pwopsd{} averages the
per-token loss within each rollout before averaging across the batch (a
per-sequence reduction), while \opsd{}'s training script accumulates the loss
in a global pool that effectively pools tokens across the batch first. Both
changes alter how single-rollout positions contribute to the gradient, and a
naive comparison conflates them. We run the full $2{\times}2$ factorial under
identical training and evaluation regimes; \Cref{tab:reduction-ablation}
reports Avg@12 on AIME 2024.

\begin{table}[h]
\centering
\small
\caption{$2{\times}2$ ablation on Qwen3-4B AIME 2024: reduction (uniform
vs.\ per-sequence) by positioning (none vs.\ position-weighted). Avg@12
mean $\pm$ sample standard deviation across three evaluation seeds. Bold
marks the column maximum. The diagonal ($\opsd{}$ and \pwopsd{}~Moderate)
reports the same evaluation runs as the corresponding rows of
\Cref{tab:main-4b}; small differences in the standard-deviation digits
reflect the use of sample (Bessel-corrected) standard deviation here. The
position-weighted rows fix the schedule to Moderate
$(w_{\min},\tau,s)=(0.25,0.30,0.10)$.}
\label{tab:reduction-ablation}
\begin{tabular*}{0.95\linewidth}{@{\extracolsep{\fill}}lc@{}}
\toprule
Configuration & AIME 2024 \\
\midrule
Uniform / global ($\opsd{}$)               & $75.19 \pm 0.42$ \\
Uniform / per-seq                          & $74.72 \pm 0.96$ \\
Pos-w / global                             & $74.54 \pm 0.64$ \\
Pos-w / per-seq (\pwopsd{}, \textbf{ours}) & $\mathbf{76.20 \pm 0.58}$ \\
\bottomrule
\end{tabular*}
\end{table}

Only the joint configuration (row 4) matches the AIME 2024 lead of
\Cref{tab:main-4b}; switching either axis alone underperforms by
$\sim 1.5$ pp. The two axes are complementary on AIME 2024 rather
than independently sufficient.

\end{document}